\documentclass[letterpaper]{article}
\pdfoutput=1
\usepackage{aaai24}
\usepackage{times}
\usepackage{helvet}
\usepackage{courier}
\usepackage[hyphens]{url}
\usepackage{graphicx}
\usepackage{amssymb}
\usepackage{caption}
\usepackage{amsmath}
\usepackage{algorithm}
\usepackage{newfloat}
\usepackage{listings}
\usepackage{algpseudocode}
\usepackage{graphicx}

\begin{document}
\twocolumn[
\begin{@twocolumnfalse}
    \begin{center}
        {\LARGE \textbf{Survey on Fundamental Deep Learning 3D Reconstruction Techniques}}\\
        \vspace{0.5cm}
        Yonge Bai\textsuperscript{1}, 
        LikHang Wong\textsuperscript{2}, 
        TszYin Twan\textsuperscript{2}\\             
        \vspace{0.5cm}
        \textsuperscript{1}McMaster University\\
        \textsuperscript{2}City University of Hong Kong\\
        \vspace{0.5cm}
        \textit{baiy58@mcmaster.ca, klhwong3-c@my.cityu.edu.hk, tytwan2-c@my.cityu.edu.hk}\\
        \vspace{0.5cm}
        \textit{July 10, 2024}
    \end{center}
    \vspace{0.5cm}
\end{@twocolumnfalse}
]

\section{Abstract}

This survey aims to investigate fundamental deep learning (DL) based 3D reconstruction techniques that produce photo-realistic 3D models and scenes, highlighting Neural Radiance Fields (NeRFs), Latent Diffusion Models (LDM), and 3D Gaussian Splatting. We dissect the underlying algorithms, evaluate their strengths and tradeoffs, and project future research trajectories in this rapidly evolving field. We provide a comprehensive overview of the fundamental in DL-driven 3D scene reconstruction, offering insights into their potential applications and limitations.
\section{Background}

3D reconstruction is a process aimed at creating volumetric surfaces from image and/or video data. This area of research has gained immense traction in recent months and finds applications in numerous domains, including virtual reality, augmented reality, autonomous driving, and robotics. Deep learning has emerged to the forefront of 3D reconstruction techniques and has demonstrated impressive results enhancing realism and accuracy.
\section{Neural Radiance Fields}
Neural Radiance Field (NeRF) is a method for novel view synthesis of complex scenes using a set of input perspectives and optimizes a model to approximate a continuous volumetric scene or surface\cite{mildenhall2020nerf}. The method represents the volume using a multilayer preceptron (MLP) whose input is a 5D vector $(x, y, z, \theta,\phi)$. $(x, y, z)$ representing the spatial location and $(\theta, \phi)$ representing the viewing direction, with an output of a 4D vector ($R,G,B,\sigma)$ representing the RGB color and a volume density. NeRFs achieved SOTA results on quantitative benchmarks as well as qualitative tests on neural rendering and view synthesis. 

\subsection{Prior Work}
NeRFs build upon prior work in RGB-alpha volume rendering for view-synthesis and the use of neural networks (NN) as implicit continuous shape representations.

\subsubsection{Volume Rendering for View-Synthesis}
This process involves, using a set of images to learn a 3D discrete volume representation, the model estimate the volume density and emitted color at each point in the 3D space, which is then used to synthesize images from various viewpoints. Prior methods include Soft 3D, which implements a soft 3D representation of the scene by using traditional stereo methods, this representation is used directly to model ray visibility and occlusion during view-synthesis \cite{Soft3DReconstruction}. Along with deep learning methods such as Neural Volumes which uses a an encoder-decoder network that transforms the input images into a 3D voxel grid, used to generate new views \cite{Lombardi:2019}. While these volumetric representation are easy to optimize by being trained on how well they render the ground truth views, but as the resolution or complexity of the scene increases the compute and memory needed to store these discretized representations become unpractical.  

\subsubsection{Neural Networks as Shape Representations}
This field of study aims to implicitly represent the 3D surface with a NN's weights. In contrast to the volumetric approach this representation encodes a description of a 3D surface at infinite resolution without excessive memory footprint as described here\cite{mescheder2019occupancy}. The NN encodes the 3D surface by learning to map a point in space to a property of that point in the 3D space, for example occupancy \cite{mescheder2019occupancy} or signed distance fields \cite{park2019deepsdf}. While this approach saves significant memory it is harder to optimize, leading to poor synthetic views compared to the discrete representations.

\section{Approach: NeRF}
NeRFs combine these two approaches by representing the scene in the weights of an MLP but view synthesis is trained using the techniques in traditional volume rendering.

\begin{figure}[t]
\centering
\includegraphics[width=\linewidth, keepaspectratio]{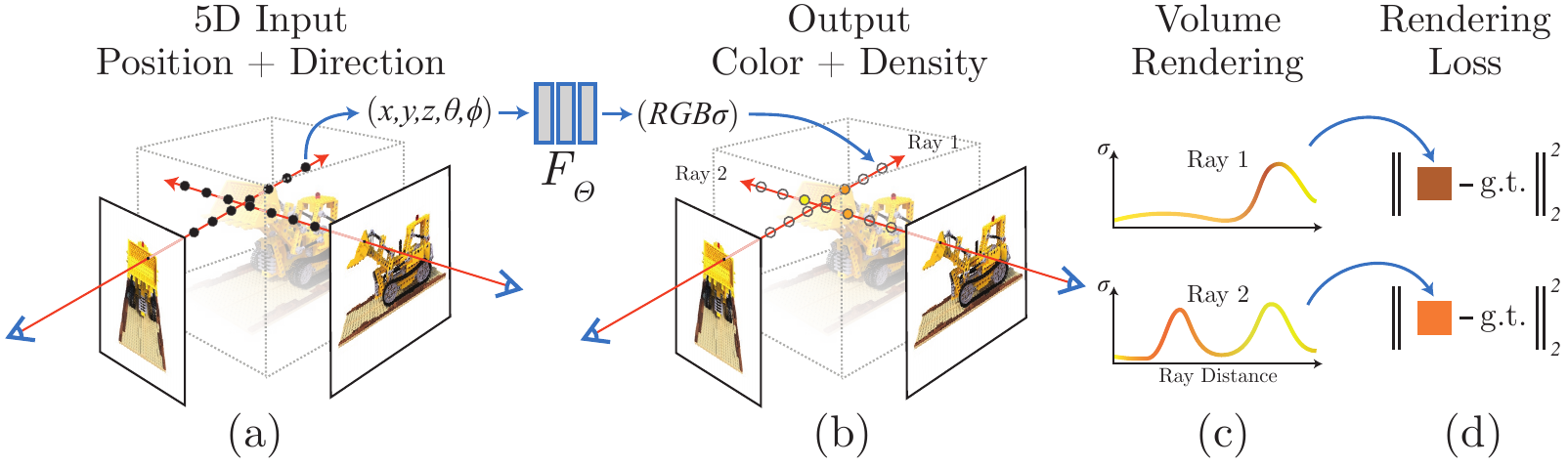}
\caption{
An overview of the neural radiance field scene representation and differentiable rendering procedure. Synthesize images by sampling 5D coordinates (location and viewing direction) along camera rays (a), feeding those locations into an MLP to predict a color and volume density (b), and using volume rendering techniques to composite these values into an image (c). This rendering function is differentiable, so we can optimize our scene representation by minimizing the residual between synthesized color and ground truth of the actual color(d).
}
\label{fig:pipeline}
\end{figure}

\subsection{Neural Radiance Field Scene Representation}
The scene is represented by 5D vector comprised of $\mathbf{x} = (x, y, z)$ and $\mathbf{d} = (\theta,\phi)$. This continuous 5D scene representation is approximated by a MLP network $F_{\Theta} : (\mathbf{x}, \mathbf{d}) \to (\mathbf{c},\sigma)$, whose weights $\Theta$ are optimized to predict each 5D input's $\mathbf{c}=(R,G,B)$ representing RGB color 
and $\sigma$ representing density. Density can be thought of as occlusion, points with a high occlusion having a higher $\sigma$ value than points with lower occlusion.

\begin{figure}[t]
\centering
\includegraphics[width=\linewidth]{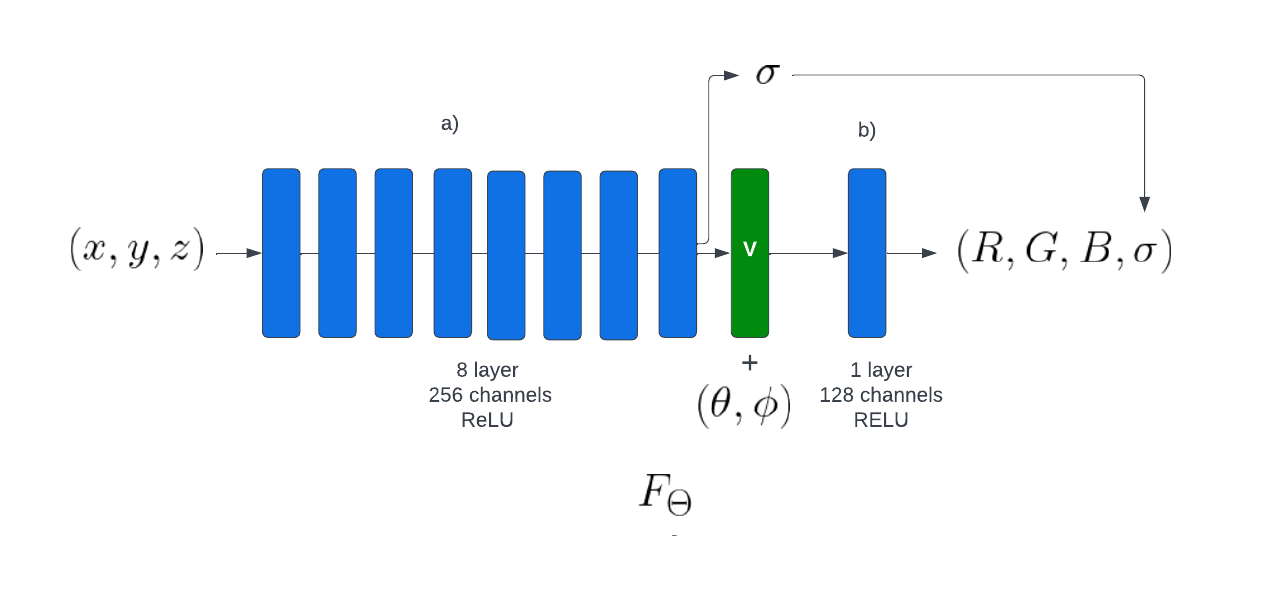}
\caption{
An overview of the NeRF model. $\mathbf{x}$ is passed into the first 8 layers, which output $\mathbf{v}$ and $\sigma$ a). $\mathbf{v}$ is concatenated with $\mathbf{d}$ and passed into the last layer, which outputs $\mathbf{c}$ b).
}
\label{fig:model}
\end{figure}

The implicit representation is held consist by forcing the network to predict $\sigma$ only as a function of $\mathbf{x}$, as density should not change as a result of viewing angle. While $\mathbf{c}$ is trained as a function of both $\mathbf{x}$ and $\mathbf{d}$. 
The MLP $F_{\Theta}$ has 9 fully-connected layers using ReLU activation functions and 256 channels per layer for the first 8 layers and 128 channels for the last layer. $F_{\Theta}$ first processes $\mathbf{x}$ with the first 8 layers outputting $\sigma$ and a 256-dimensional feature vector $\mathbf{v}$. {$\mathbf{v}$} is then concatenated with $\mathbf{d}$ and passed into the final layer that outputs $\mathbf{c}$. This process is shown in Figure  \ref{fig:model}. 

\subsection{Volume Rendering with Radiance Fields}
The color of any ray passing through the scene is rendered using principles from classical volume rendering. 

\begin{equation}
\label{eq:volume_render}    
    \hat{C}(\mathbf{r})=\sum_{i=1}^{N}w_{i}c_{i}, \text{ where } w_{i} = T_{i}\alpha_{i}
\end{equation} 

Equation \eqref{eq:volume_render} can be explained as the color $c_{i}$ of each point being weighted by $w_{i}$. $w_{i}$ is made up of $T_{i}=\exp(- \sum_{j=1}^{i-1} \sigma_{i} \delta_{i})$ where $\sigma_{i}$ is the density and $\delta_{i}$ is the distance between adjacently sampled points. 
$T_{i}$ denotes the accumulated transmittance until point $i$ which can be thought of as the amount of light blocked earlier along the ray, and $\alpha_{i} = 1-\exp(-\sigma_{i}\delta_{i})$ denoting the opacity at point $i$. Thus the color predicted at point with higher transmittance and opacity (the beginning of surfaces) contribute more to final predicted color of ray $\mathbf{r}$.

\subsection{Optimizing a NeRF}
The previous sections covered the core components to NeRFs but the original paper had two more techniques to achieve SOTA quality---positional encoding and hierarchical volume sampling.

\subsubsection{Positional Encoding}
The authors found that directly feeding in $(x,y,z,\theta,\phi)$ to $F_{\Theta}$ resulted in poor performance. As a result, they chose to map the inputs to a higher dimensional space using high frequency functions, this enabled the model to better fit data with high variations.
Thus $F_{\Theta}$ is reformulated as a composition of two functions $F_{\Theta} = F_{\Theta}'\circ\gamma$. $F_{\Theta}'$ being the original MLP and $\gamma$ defined as:

\begin{equation}
    \gamma(x) = \left(
    \begin{array}{cc}
    \sin(2^0 \pi x), & \cos(2^0 \pi x) \\
    \vdots & \vdots \\
    \sin(2^{L-1} \pi x), & \cos(2^{L-1} \pi x)
    \end{array} \right)
\end{equation}

$\gamma(\cdot)$ is applied to $(x,y,z)$ in $\mathbf{x}$ with $L = 10$ and $(\theta,\phi)$ with $L = 4$.

\begin{figure}[t]
\centering
\includegraphics[width=\linewidth]{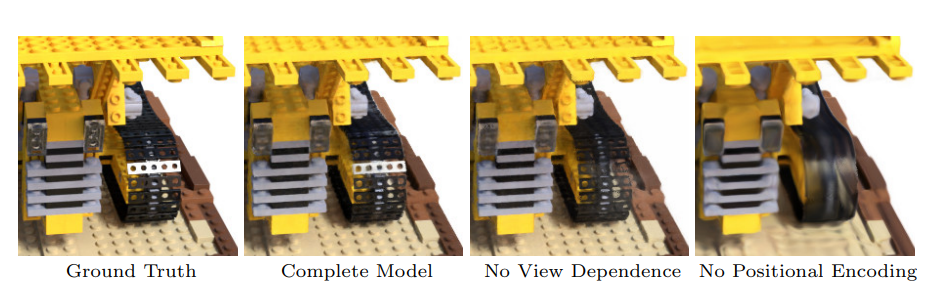}
\caption{
Visualizing how the model improves the positional encoding. Without it the model is unable to represent high variation geometries and textures resulting in an over smoothed, blurred appearance. Also how removing view dependency affect the models ability to render lighting and reflections.
}
\label{fig:positional_encodings}
\end{figure}

$\gamma$ is a mapping from $\mathbb{R}$ to $\mathbb{R}^{2L}$ that significantly improves performance (Figure \ref{fig:positional_encodings}).

\subsubsection{Hierarchical Volume Sampling}
\begin{figure}[t]
\centering
\includegraphics[width=\linewidth]{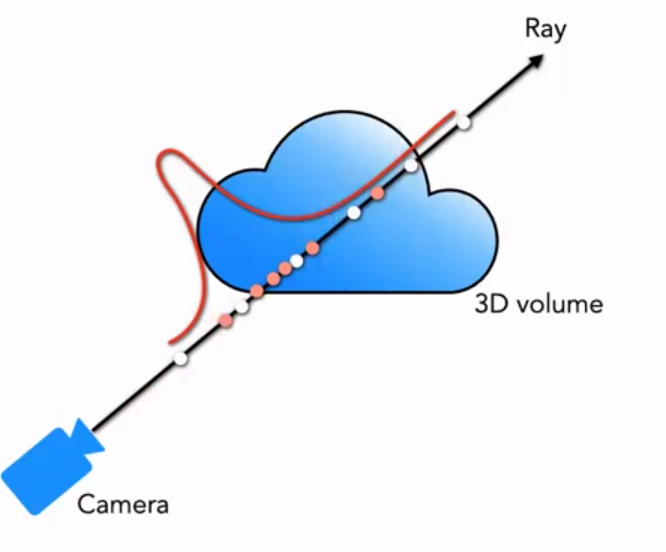}
\caption{ 
Illustrating hierarchical sampling, where samples are proportional to their contribution to the final volume render. 
}
\label{fig:hierarchical_sampling}
\end{figure}

Free space and occluded region contribute much less to the quality of the NeRF compared to areas at the beginning of a surface, but with uniform sampling, are sampled at the same rate. So the authors proposed a hierarchical representation that increases rendering efficiency and quality by allocating samples proportional to their expected effect shown in \ref{fig:hierarchical_sampling}. For example, if the object in question was a ball, there would be less samples taken in the open space in front of the ball and inside of the ball verses samples directly on the ball's surface.

This is done by optimizing two networks. One "coarse" and one "fine". The course network samples points uniformly along the ray, while the fine network is biases toward the relevant part of the volume by normalizing the per sample weights $w_{i}$ described in equation \eqref{eq:volume_render}, this allows one to treat the weight of each point as a probability distribution which is sampled to train the fine network. This procedure allocates more samples to regions expected to contain visible content.

\subsection{Limitations}
While having groundbreaking abilities to render photorealistic 3D volumes from 2D images, the original NeRF methodology suffered from several limitations. These limitations include:

\subsubsection{Computational Efficiency}
The optimization of a single scene took 100-300k iterations to converge of a single NVIDIA V100 GPU which corresponded to 1-2 days \cite{mildenhall2020nerf}. This poor computational efficiency is a product of dense sampling of rays for rendering. This dense sampling approach helped in capturing fine details and accurately representing complex scenes, but it significantly increases the computational load.

\subsubsection{Lack of Generalizability}
NeRFs are inheritably inflexible due to the as models overfit to one scene. A NeRF cannot be adapted for novel scenes without complete retraining.

\subsubsection{Difficulty of Editing}
Modifying content in NeRFs such as moving or removing object is very difficult. Since the model represents the scene as a continuous function and does not store geometric information.

\subsubsection{Data requirements}
NeRFs require a lot of data to produce high quality results show in the original paper. 
The synthetic 3D models as lego bulldozer and pirate ship took about 100 image and the real life scenes such as the flower and conference room each requiring around 60 \cite{mildenhall2020nerf}.

\subsection{Transient Artifacts}
\begin{figure}[t]
\centering
\includegraphics[width=\linewidth]{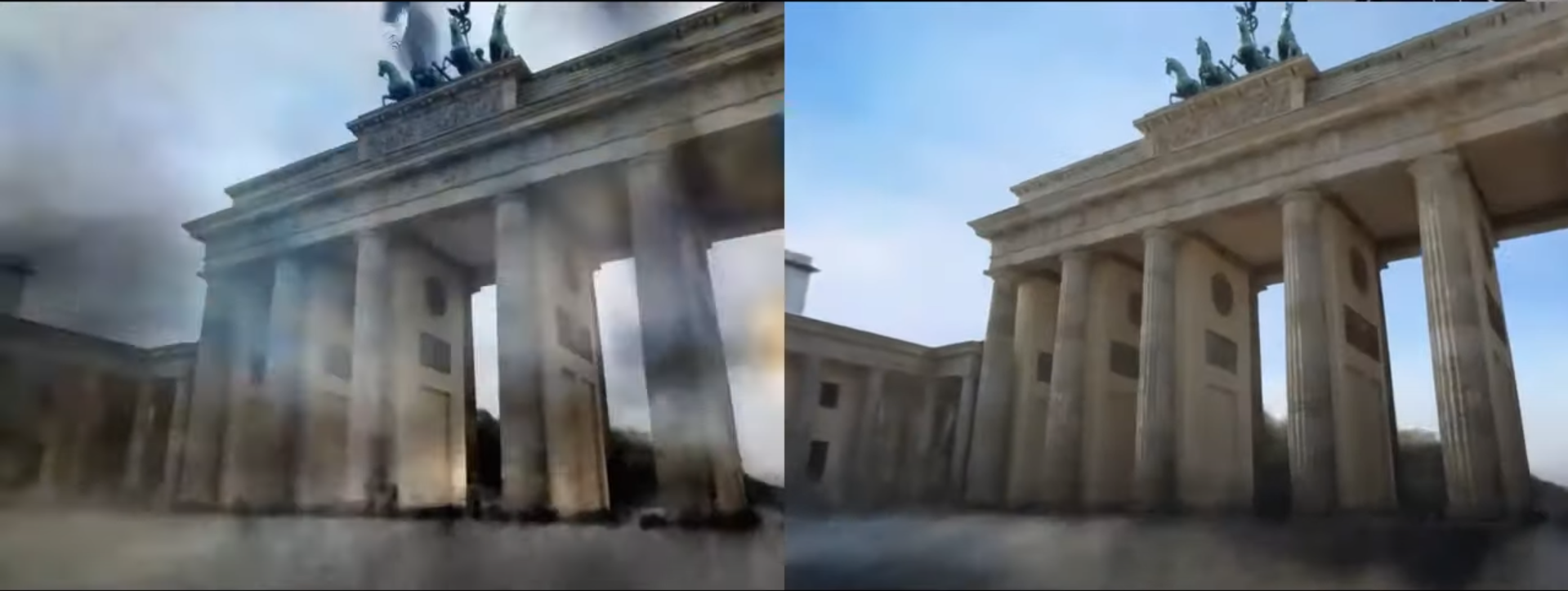}
\caption{ 
Comparison made in the paper NeRF in the Wild \cite{martinbrualla2021nerf}, where the original NeRF (left) noisy artifacts compared to NeRF-W (right).
}
\label{fig:nerf_arifacts}
\end{figure}

The original NeRFs assume that the world is geometrically, materially, and photometrically static. Therefore requiring that any two photographs taken at the same position and orientation must be identical \cite{martinbrualla2021nerf} they do not have a way to adjust for transient occlusions or variable appearance which result in artifacts and noise when this assumption fails such as with real world images. This is clearly show in \ref{fig:nerf_arifacts}.

\section{Instant-NGP}
\subsection{Overview}
Instant-NGP\cite{M_ller_2022}, proposed by Nvlabs, is a method that significantly reduce the computation demand of original NeRFs. It leverages multi-resolution hash grids to improve memory usage and optimizes 3D reconstruction performance.
\subsection{Prior work}
\subsubsection{Learnable positional encoding}
Learnable positional encoding refers to positional encodings that are parameterized for specific positions in a continuous 3D space. The positional encoding for a point $p$ in 3D space can be represented as:

\[
\mathbf{pe}(p) = \sigma(\mathbf{W} \mathbf{p} + \mathbf{b})
\]

where $\mathbf{p} = (x, y, z)^T$ represents the coordinates of the position in 3D space, $\mathbf{W}$ is a learnable weight matrix, $\mathbf{b}$ is a bias vector, and $\sigma$ denotes a non-linear activation function.\\

These positional encodings can then be integrated into a neural network model to facilitate the learning of spatial relationships.

\begin{figure}[h]
    \centering
    \includegraphics[width=1\linewidth]{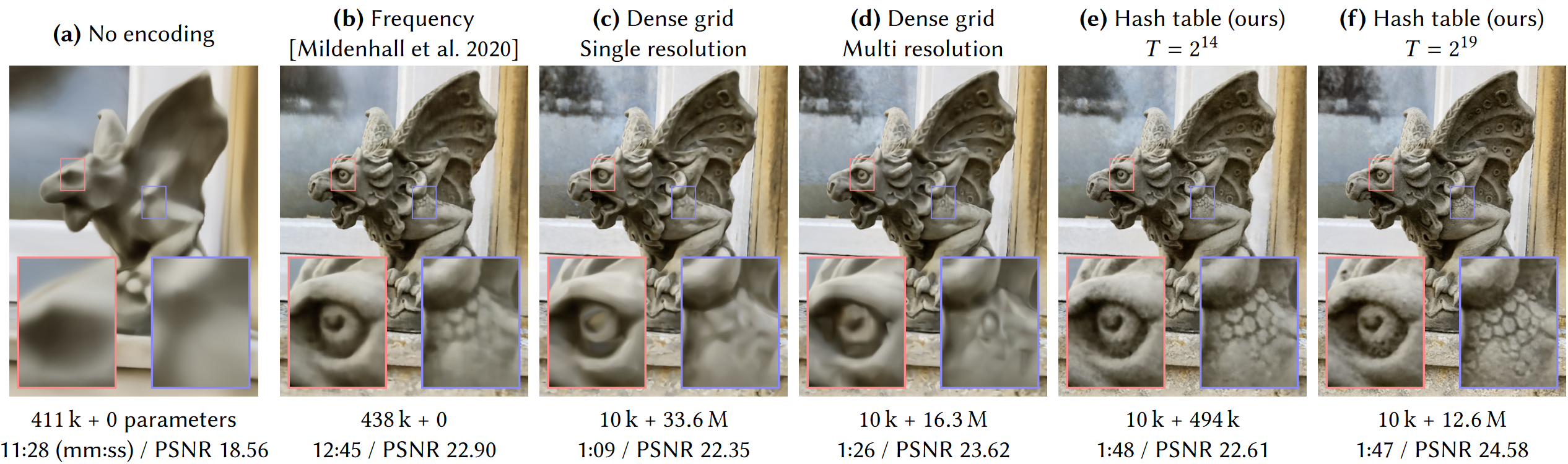}
    \caption{Experiment performed in the original paper. as the number of parameters used for learning the positional encoding increases, the image becomes clearer and sharper.}
    \label{fig:enter-label}
\end{figure}

\subsection{Algorithm}

\subsubsection{Multi-Resolution Hash Encoding}

\begin{figure}[h]
    \centering
    \includegraphics[width=1\linewidth]{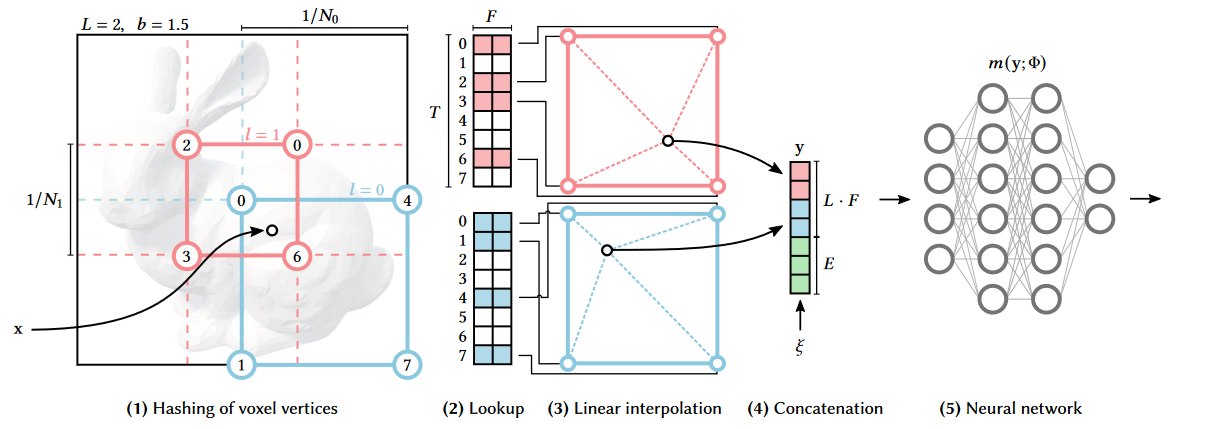}
    \caption{Illustration of the multiresolution hash encoding represented in 2D in the original paper.}
    \label{fig:mhei}
\end{figure}

One of the key components of Instant-NGP (Instant Neural Graphics Primitives) is the Multi-Resolution Hash Encoding. Instead of learning the positional encoding for the entire 3D space, the 3D space is first scaled to fit within a normalized range of 0 to 1. This normalized space is then replicated across multiple resolutions and each subdivided into grids of varying densities.
This captures both coarse and fine details in the scene. Each level focuses on learning the positional encodings at the vertices of the grids. Mathematically, this can be expressed as follows:

\[
\mathbf{p}_{scaled} = \mathbf{p} \cdot \mathbf{s}
\]

where \(\mathbf{p}\) represents the original coordinates in 3D space, and \(\mathbf{s}\) is a scaling factor that normalizes the space to the [0, 1] range. Following the scaling, the coordinates are hashed into a multi-resolution structure using a spatial hash function:

\[
h(\mathbf{x}) = \left( \bigoplus_{i=1}^d (x_i \cdot \pi_i) \right) \mod T
\]

Here, \(d\) is the dimensionality of the space (e.g., 3 for 3D coordinates), \(\mathbf{x} = (x_1, x_2, \dots, x_d)\) represents the scaled coordinates, \(\bigoplus\) denotes the bit-wise XOR operation, \(\pi_i\) are large prime numbers unique to each dimension, and \(T\) is the size of the hash table. This function maps spatial coordinates to indices in the hash table, where the neural network's parameters are stored or retrieved, linking specific spatial locations to neural network parameters.
\\ \\
In a nutshell, a hash table is assigned to each level of resolution. For each resolution, each vertex is mapped to a entry in the resolution's hash table. Higher-resolution have larger hash tables compared to lower-resolutions. Every resolution's hash table map each of it's vertices to an individual set of parameters that learn their positional encodings.

\subsubsection{Learning positional encoding}
During training, when the model is exposed to images from different viewpoints, the NN adjusts the parameters stored in the hash table to minimize the difference between the rendered images and the actual training images. The loss $L$ can be expressed as:
\[
L(\theta) = \frac{1}{2} \sum_{i=1}^m (R(x_i, \theta) - y_i)^2
\]

where $R(x_i, \theta)$ is the rendered image based on parameters $\theta$ and view point $x_i$, $y_i$ is the corresponding actual image, and $m$ is the number of pixels or data points considered. We can then learn these parameters using different optimization techniques like gradient descent.

\subsubsection{Hash Collisions}
Hash collisions are avoided by assigning a hash table at each resolution that long enough to ensure one-to-one mapping from entries to positional encodings. 

\subsection{Performance}
\begin{figure}[h]
    \centering
    \includegraphics[width=1\linewidth]{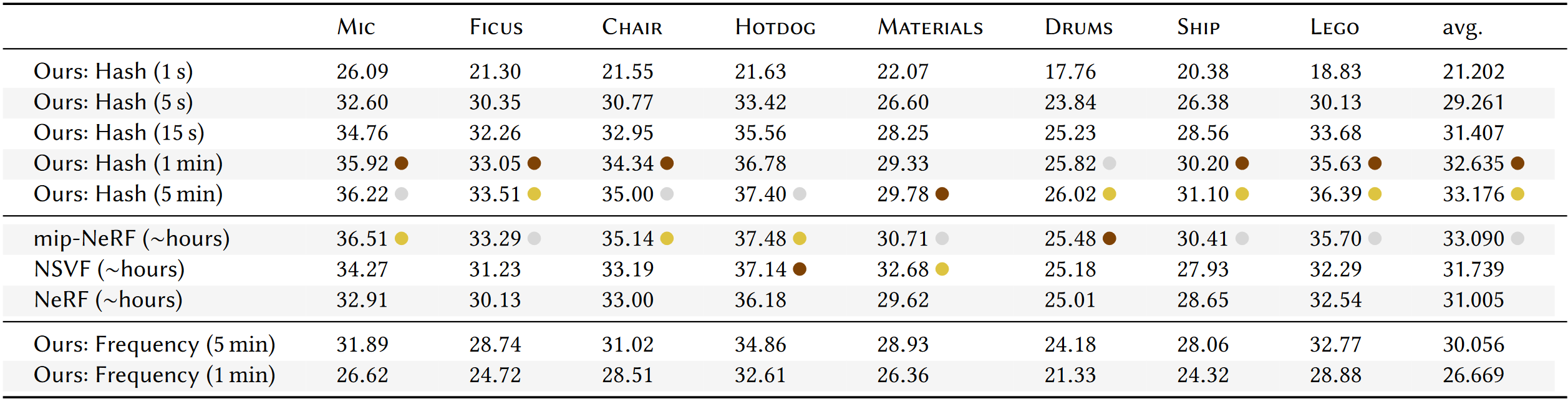}
    \caption{The figure adapted from the original paper compares the Peak Signal to Noise Ratio (PSNR) performance of various NeRF implementations, including the author's multi-resolution hash encoding method, against other models that require hours of training. First row is the name of the object constructed.}
    \label{fig: tabingp}
\end{figure}

As shown in figure \ref{fig: tabingp}, Instant-NGP achieved a notable 20-60× speed improvement compared to compared to the original NeRFs while maintaining it's quality.

\subsection{Limitations}

Instant-NGP focuses on speeding up the computation and training processes of NeRFs. However, it still suffers from many of the same issues such as generalability different datasets or unseen scenarios. In the next section, we introduce LDM based techniques for 3D reconstruction to address the issue of generalizable.
\section{Latent-Diffusion-Model based 3D reconstruction}
\subsection{Background}
\begin{figure}[H]
    \centering
    \includegraphics[width=1\linewidth]{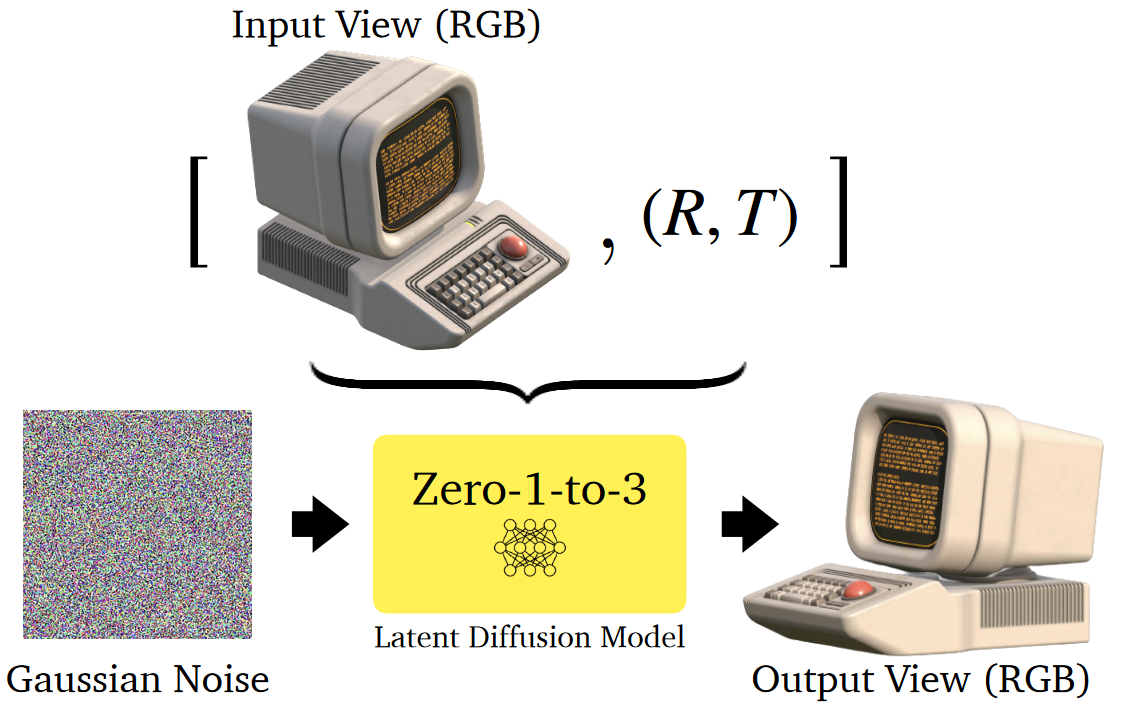}
    \caption{High Level Picture of Zero-1-to-3 from the original paper}
    \label{fig:zero-1-2-3}
\end{figure}

Traditional 3D-Reconstruction algorithms rely heavily on the training data to capture all aspects of the volume. Humans, however, are able to estimate a 3D surface from a single image. This concept is the foundation of the Zero-1-to-3\cite{liu2023zero1to3} framework developed out of Columbia University which introduces a diffusion-based 3D reconstruction method. Zero-1-to-3 utilizes a LDM, originally designed for text-conditioned image generation, to generate new perspectives of an image based on a camera's extrinsic parameters like rotation and translation. Zero-1-to-3 leverages the geometric priors learned by large-scale LDMs, allowing the generation of novel views from a single image. Zero-1-to-3 demonstrates strong zero-shot generalization capabilities, outperforming prior models in both single-view 3D reconstruction and novel view synthesis tasks. See Figure \ref{fig:zero-1-2-3}. 
\subsection{Prior Work: Denoising Diffusion Probabilistic Models}

Denoising Diffusion Probabilistic Models (DDPMs)\cite{ho2020denoising} are a class of generative models that transform data by gradually adding noise over a sequence of steps, then learning to reverse this process to generate new samples from noise.

\subsubsection{Forward Process}
The forward process in DDPM is a Markov chain that gradually adds Gaussian noise to the data over \( T \) timesteps. The process can be mathematically described as:

\[
x_{t} = \sqrt{\alpha_t} x_{t-1} + \sqrt{1 - \alpha_t} \epsilon, \quad \epsilon \sim \mathcal{N}(0, I)
\]

where \( x_0 \) is the original data, \( x_t \) is the data at timestep \( t \), \( \epsilon \) is the noise, and \( \alpha_t \) is the variance schedule parameters that determine how much noise is added at each step. 

\begin{figure}[h]
    \centering
    \includegraphics[width=1\linewidth]{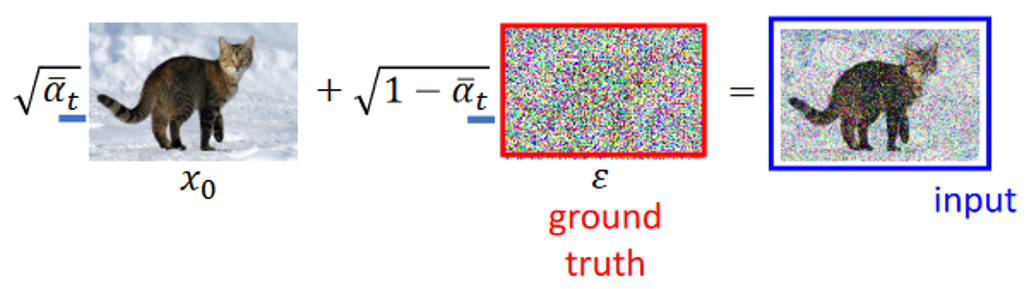}
    \caption{Forward Process of DDPM. Adapted from \cite{lee_ddpm_video}.}
    \label{fig:ddpm_forward}
\end{figure}
\subsubsection{Reverse Process}

The reverse process aims to reconstruct the original data from the noise by learning a parameterized model \( p_{\theta} \). The reverse process is also a Markov chain, described as:

\[
x_{t-1} = \frac{1}{\sqrt{\alpha_t}} \left( x_t - \frac{1 - \alpha_t}{\sqrt{1 - \bar{\alpha}_t}} \epsilon_{\theta}(x_t, t) \right) + \sigma_t z, \quad z \sim \mathcal{N}(0, I)
\]

where \( \epsilon_{\theta}(x_t, t) \) is a neural network predicting the noise, \( \sigma_t \) is the standard deviation of the reverse process noise, and \( \bar{\alpha}_t = \prod_{s=1}^t \alpha_s \) is the cumulative product of the \( \alpha_t \) values.

\subsubsection{Training}

The training of DDPMs involves optimizing the parameters \( \theta \) of the neural network to minimize the difference between the noise predicted by the model and the actual noise added during the forward process. The loss function is typically the mean squared error between these two noise terms:

\[
L(\theta) = \mathbb{E}_{t, x_0, \epsilon}\left[ \| \epsilon - \epsilon_{\theta}(x_t, t) \|_2^2 \right]
\]

where \( x_t \) is computed during the forward process and \( \epsilon \) is the Gaussian noise added at each step.

\subsubsection{Sampling}

To generate new samples, the reverse process is initialized with pure noise \( x_T \sim \mathcal{N}(0, I) \) and iteratively applies the reverse steps to produce samples approximating the distribution of the original data \( x_0 \).

\begin{algorithm}
\caption{DDPM Sampling Algorithm}
\begin{algorithmic}[1] 
\Procedure{DdpmSampling}{$\theta, T, \{\alpha_t\}$}
    \State \textbf{Input:} Trained model parameters $\theta$, total timesteps $T$, noise schedule $\{\alpha_t\}$
    \State \textbf{Output:} A sample approximating the data distribution

    \State \textbf{Initialize:} Draw $x_T \sim \mathcal{N}(0, I)$ {Start with pure noise}
    \For{$t = T$ \textbf{down to} 1}
        \State Calculate $\bar{\alpha}_t = \prod_{s=1}^t \alpha_s$
        \State Calculate $\sigma_t^2 = \frac{1-\bar{\alpha}_{t-1}}{1-\bar{\alpha}_t} \cdot (1-\alpha_t)$
        \State Predict noise $\epsilon_t = \epsilon_\theta(x_t, t)$
        \If{$t > 1$}
            \State $x_{t-1} = \frac{1}{\sqrt{\alpha_t}} \left( x_t - \frac{1 - \alpha_t}{\sqrt{1 - \bar{\alpha}_t}} \epsilon_t \right) + \sigma_t z, \quad z \sim \mathcal{N}(0, I)$
        \Else
            \State $x_0 = \frac{1}{\sqrt{\alpha_t}} \left( x_t - \frac{1 - \alpha_t}{\sqrt{1 - \bar{\alpha}_t}} \epsilon_t \right)$ {Final denoising step}
        \EndIf
    \EndFor
    \State \textbf{Return} $x_0$
\EndProcedure
\end{algorithmic}
\end{algorithm}

\begin{figure}[h]
    \centering
    \includegraphics[width=1\linewidth]{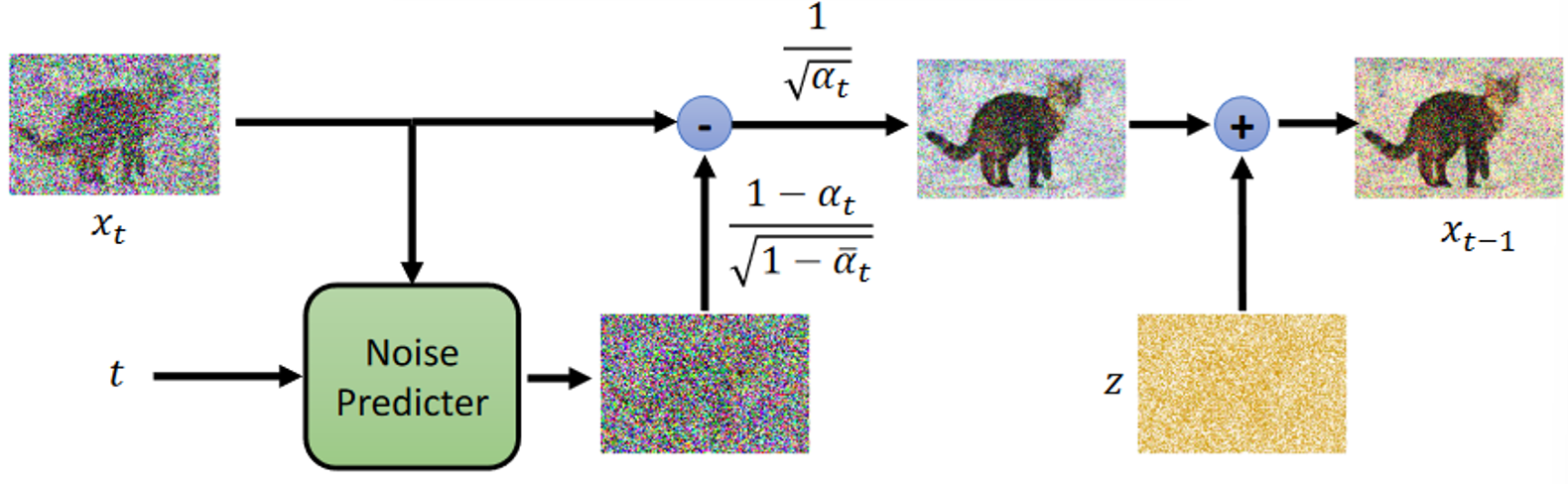}
    \caption{Sampling Process. Adapted from \cite{lee_ddpm_video}.}
    \label{fig:enter-label}
\end{figure}
\subsection{Latent Diffusion Model in Zero-1-to-3}
Latent Diffusion Models\cite{rombach2022highresolution} proposed in 2021 are a type of generative model that combines the strengths of diffusion models and Variationsal Autoencoders(VAEs). Traditional DDPMs operates in the image pixel space, which requires more computation. LDMs compress the full image data space into a latent space before the diffusion and denosing process, improving efficiency and scalability in generating high-quality images.
\begin{figure}[H]
    \centering
    \includegraphics[width=\linewidth]{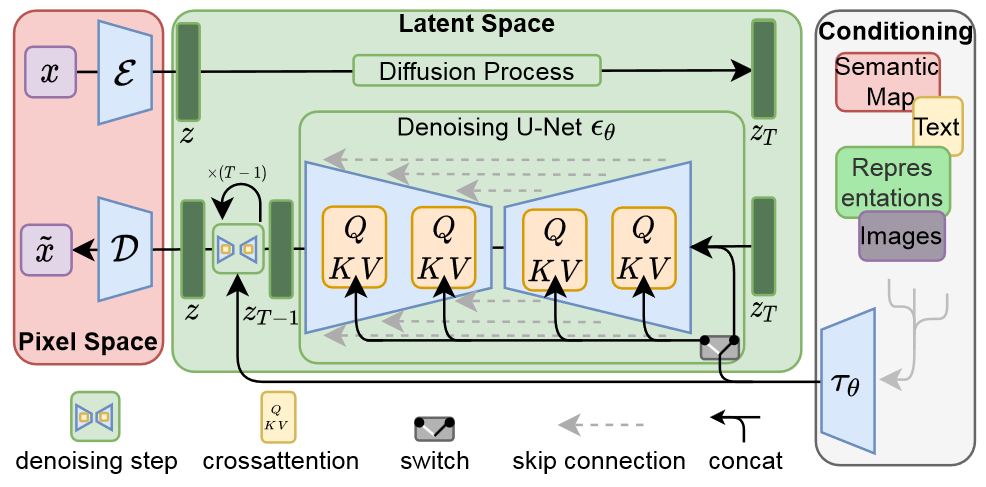}
    \caption{The architecture of Latent Diffusion Model in the original paper.}
    \label{fig:enter-label}
\end{figure}
\subsubsection{Training Latent Diffusion Models $\epsilon_{\theta}$}
LDM is trained in two main stages. First, a VAE is used to learn an encoding function $E(x)$ and a decoding function $D(z)$, where $x$ represents the high-resolution image and $z$ is it's latent representation. The encoder compresses $x$ to $z$, and the decoder attempts to reconstruct $x$ from $z$. 

\subsubsection{Training VAE} The VAE optimizes the parameters $\phi$ (encoder) and $\psi$ (decoder) by minimizing the reconstruction loss combined with the KL-divergence loss:
\begin{figure}[h]
    \centering
    \includegraphics[width=\linewidth]{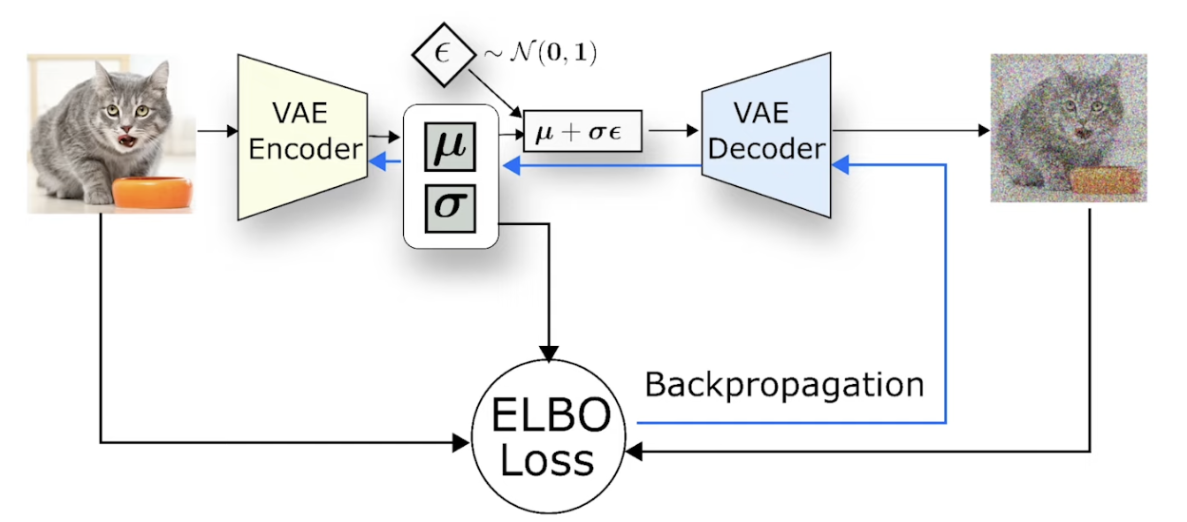}
    \caption{Training VAE. Figure from Lightning AI}
    \label{fig:training_VAE}
\end{figure}
\begin{equation}
    \mathcal{L}_{VAE}(\phi, \psi) = \mathbb{E}_{q_\phi(z|x)}[\log p_\psi(x|z)] - D_{KL}(q_\phi(z|x) \| p(z))
\end{equation}

\subsubsection{Training Attention-U-Net Denoiser} 
\begin{figure}[h]
    \centering
    \includegraphics[width=1\linewidth]{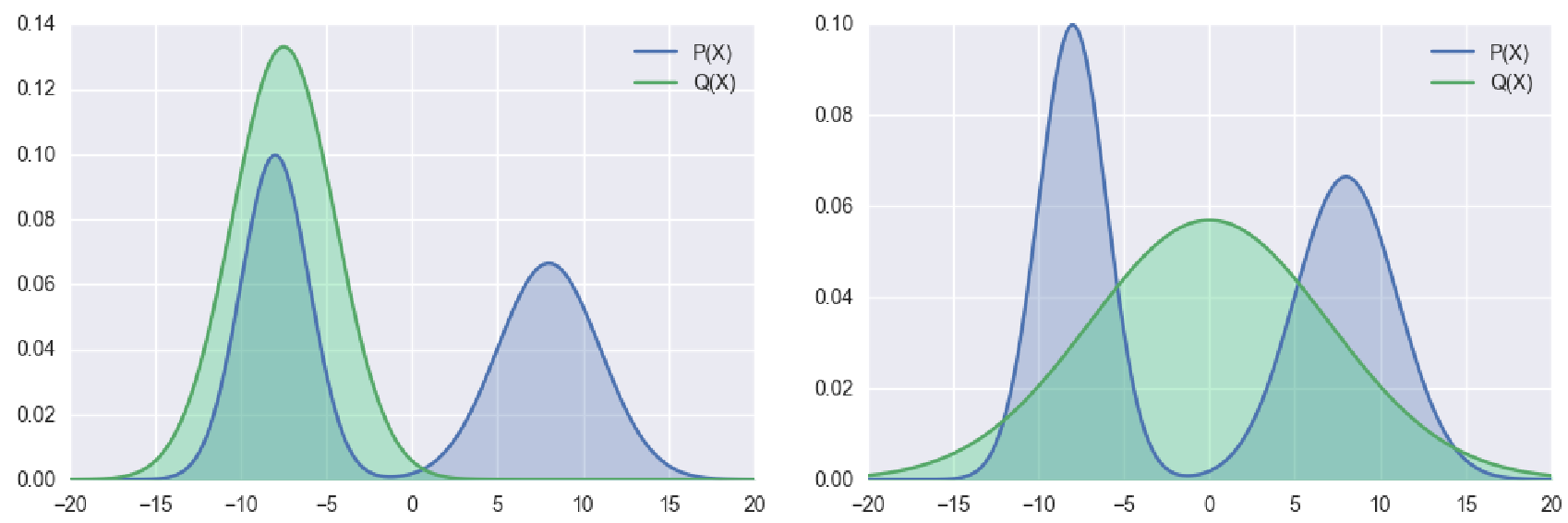}
    \caption{On the left (bad) example, the difference between $Q(x)$ distribution and $P(x)$ is not minimized. On the right (good) example, the difference between $Q(x)$ distribution and $P(x)$ is minimized. (from \cite{Kristiadi2016}) }
    \label{fig:min_dis}
\end{figure}

In the second stage, an Attention-U-Net is trained as the denoising model in the latent space. This model learns a sequence of denoising steps that transform a sample from a noise distribution $p(z_T)$ to the data distribution $p(z_0)$ over T timesteps. The U-Net model parameter $\theta$ are optimized by minimizing the expected reverse KL-divergence between the true data distribution and the model distribution as follows:
\begin{equation}
    \mathcal{L}(\theta) = \mathbb{E}_{z_0, \epsilon \sim \mathcal{N}(0, I), t} \left[ \|\epsilon - \epsilon_\theta(z_t, t)\|^2 \right]
\end{equation}
where $z_t = \sqrt{\bar{\alpha}_t} z_0 + \sqrt{1 - \bar{\alpha}_t} \epsilon$, and $\bar{\alpha}_t$ is the variance schedule. We use KL-divergence to calculate "how different" the true data distribution and the model distribution are. We aim to make the latent space distribution of the model similar to that of the real world. This is the key to generating realistic images. Figure \ref{fig:min_dis} shows example of minimizing two distributions.
\begin{figure}[h]
    \centering
    \includegraphics[width=1\linewidth]{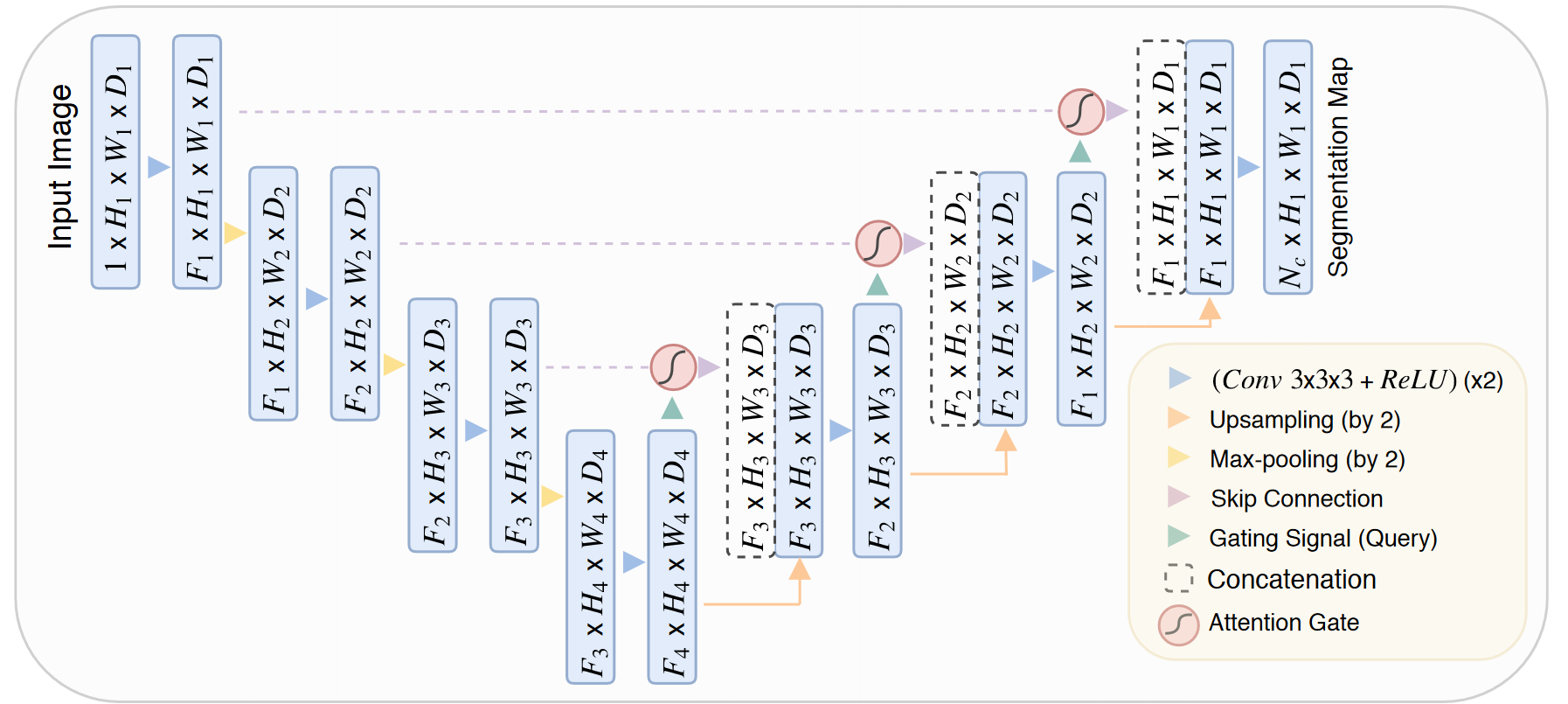}
    \caption{Attention-U-Net Architecture in the original paper}
    \label{fig:u-net}
\end{figure}
\begin{figure}[h]
    \centering
    \includegraphics[width=1\linewidth]{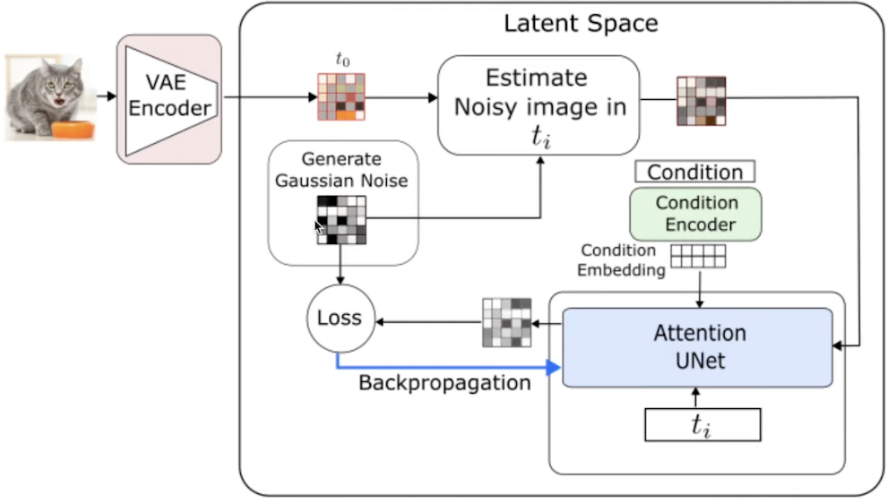}
    \caption{Training Attention-U-Net Denoiser. Figure from Lightning AI}
    \label{fig:enter-label}
\end{figure}

\subsubsection{Conditioning on Camera Parameters}
\label{subsec:conditioning}

The third stage in the Zero-1-to-3 framework focuses on the conditioning of the LDM based on camera extrinsic parameters such as rotation ($R$) and translation ($t$). This conditioning is critical for generating novel views of the object, which are essential for effective 3D reconstruction from a single image.

\paragraph{Mechanism of Conditioning}
In this stage, the previously trained latent representations are manipulated according to the desired camera transformations to simulate new perspectives. This process involves:

\begin{itemize}
    \item \textbf{Camera Transformations:} Adjusting the latent variables $z$ to reflect changes in viewpoint due to different rotations $R$ and translations $t$.
    \item \textbf{Transformation Implementation:} This could be achieved either through a learned transformation model within the LDM framework or by applying predefined transformation matrices directly to the latent vectors.
\end{itemize}

\paragraph{Training the Model for Conditional Output}
The model is further trained to handle conditional outputs effectively, which involves:

\begin{itemize}
    \item \textbf{Data Preparation:} The official code used the RTMV dataset\cite{tremblay2022rtmv} where objects are captured from multiple viewpoints to pair latent representations with corresponding camera parameters.
    \item \textbf{Model Adaptation:} Extending the latent diffusion model training to not only generate images from the latent representation $z$ but also new perspectives of the images from it's transformed latent representation $z'$.
    \item \textbf{MSE Loss:} We compute the MSE between the output image and real image with respect to $R$ and $t$. A "near-view consistency loss" that calculate the MSE between the image rendered from a view and the image rendered from a nearby view is also used to maintain the consistency in 3D reconstruction.
\end{itemize}

\paragraph{Novel View generation}
To generate a novel view, the following transformation is applied to the latent space:

\begin{equation}
z' = f(z, R, t)
\end{equation}

Where $f$ is the transformation function that modifies $z$ based on $R$ and $t$. The Zero-1-to-3 model generates the novel view as follows:

\begin{equation}
x' = D(\epsilon_\theta(z'))
\end{equation}

Where $x'$ represents the image generated from the new perspective, and $D$ is the decoder part of the LDM that synthesizes the final image output from the transformed latent representation $z'$.

\subsubsection{3D reconstruction $\epsilon_{\theta}$}
The 3D reconstruction is performed by first generating multiple views of the object using the above method for various $R$ and $t$ matrices. Each generated image $x'$ provides a different perspective of the object. These images are then used to reconstruct the 3D model:

\begin{equation}
    \text{3D Model} = \text{Integrate}(\{x'\})
\end{equation}

\begin{figure}[h]
    \centering
    \includegraphics[width=1\linewidth]{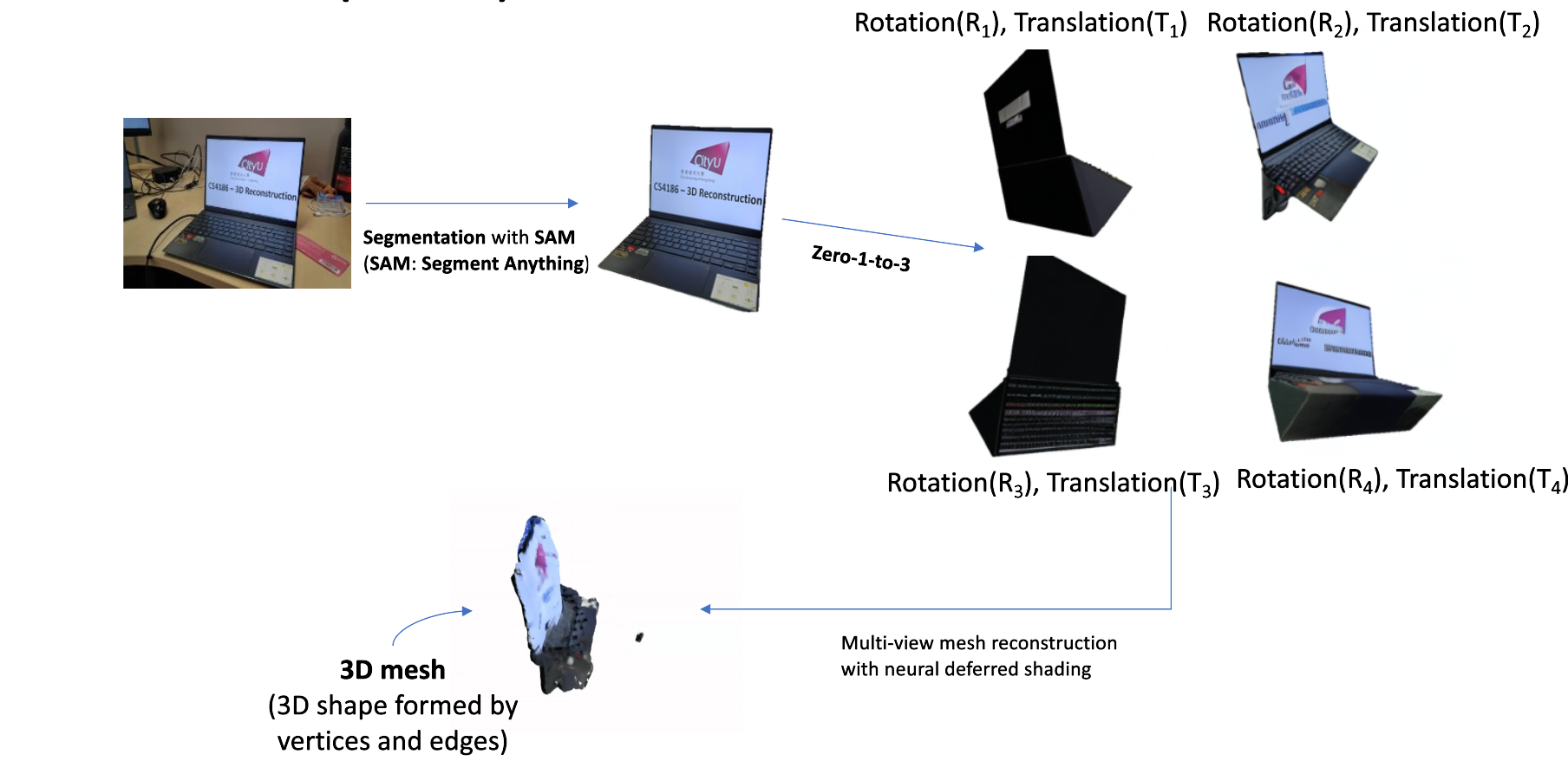}
    \caption{Demo of 3D reconstruction from Zero-1-to-3}
    \label{fig:demo}
\end{figure}

The integration process typically involves techniques like volumetric fusion or multi-view stereo algorithms, which consolidate the information from different images to create a detailed 3D representation of the object, as shown in figure \ref{fig:demo}.

\subsection{Limitations for diffusion-based and NeRF-based 3D reconstruction}
\begin{itemize}
    \item \textbf{Flexibility and Real-Time 3D scene Rendering:} Training the Zero-1-to-3 model for 3D scenes reconstruction typically require iterative denoising processes during sampling, which can be computationally intensive and slow. This makes them less suitable for applications requiring real-time rendering.
    \item \textbf{Implicit Representation Ambiguity:} Both NeRF and Diffusion models represent the 3D object implicitly; they do not explicitly construct the 3D space. NeRFs utilize the weights of an MLP to represent a 3D scene and LDMs use the latent space for new perspectives generation for 3D reconstruction. While implicit representation saves significant space it may lead to ambiguities in interpreting the model.
\end{itemize}
\section{Approach: 3D Gaussian Splatting}
\subsection{Background}
Throughout the evolution of 3D scene reconstruction, explicit representations such as meshes and point clouds have always been preferred by developers and researchers due to their clearly defined structure, fast rendering, and ease of editing. NeRF-based methods have shifted towards continuous representation of 3D scenes. While the continuous nature of these methods helps optimization, the stochastic sampling required for rendering is costly and can result in noise \cite{kerbl20233d}. On top of that the implicit representation's lack of geometric information does not lend itself well to editing.
 
\subsection{Overview}

3D Gaussian Splatting provides a high-quality novel view with real-time rendering for 3D scenes, these are achieved with the utilization of the Gaussian function to present a smooth and accurate texture using captured photos of a scene. 

    \begin{figure}[!h]
    	\includegraphics[width=\linewidth]{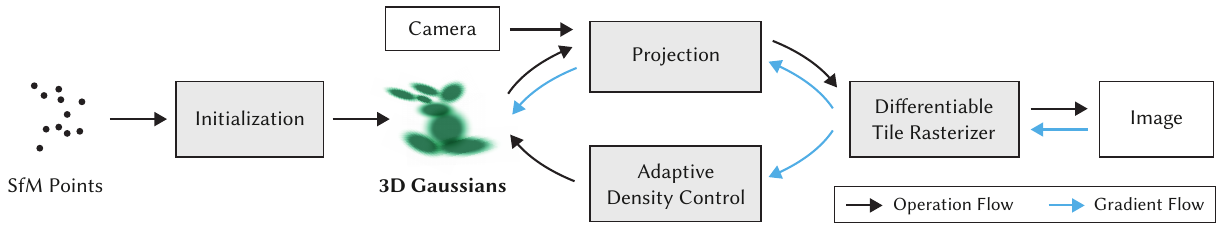}
        \label{fig:density_control}
    	\caption{
    		Overview of 3D Gaussian Splatting process.}
    \end{figure}

To reconstruct a 3D model using 3D Gaussian splatting, first, a video of the object is taken from different angles, then converted into frames representing static scenes at different camera angles. Structure from Motion (SfM) with feature detection and matching techniques such as SIFT is then applied to these images to produce a sparse point cloud. The 3D data points of the point cloud are then represented by 3D Gaussians. With the optimization process, adaptive density control of Gaussians, and high-efficiency algorithm design, realistic views of the 3D model can be reconstructed with a high frame rate. 

The algorithm of 3D Gaussian Splatting is demonstrated below, where it can be slit into 3 parts: initialization, optimization, and density control of Gaussians.

\subsection{Algorithm}
	\begin{algorithm}[!h]
		\caption{Optimization and Densification\\
		$w$, $h$: width and height of the training images}
		\label{alg:optimization}
		\begin{algorithmic}
			\State $M \gets$ SfM Points	\Comment{Positions}
			\State $i \gets 0$	\Comment{Iteration Count}
			
			\While{not converged}
			
			\State $V, \hat{I} \gets$ SampleTrainingView()	\Comment{Camera $V$ and Image}
			\State $I \gets$ Rasterize($M$, $S$, $C$, $A$, $V$)	\Comment{Alg.~\ref{alg:rasterize}}
			
			\State $L \gets Loss(I, \hat{I}) $ \Comment{Loss}
			
			\State $M$, $S$, $C$, $A$ $\gets$ Adam($\nabla L$) \Comment{Backprop \& Step}

			\If{IsRefinementIteration($i$)}
			\ForAll{Gaussians $(\mu, \Sigma, c, \alpha)$ $\textbf{in}$ $(M, S, C, A)$}
			\If{$\alpha < \epsilon$ or IsTooLarge($\mu, \Sigma)$}	\Comment{Pruning}
			\State RemoveGaussian()	
			\EndIf
			\If{$\nabla_p L > \tau_p$} \Comment{Densification}
			\If{$\|S\| > \tau_S$}	\Comment{Over-reconstruction}
			\State SplitGaussian($\mu, \Sigma, c, \alpha$)
			\Else								\Comment{Under-reconstruction}
			\State CloneGaussian($\mu, \Sigma, c, \alpha$)
			\EndIf	
			\EndIf
			\EndFor		
			\EndIf
			\State $i \gets i+1$
			\EndWhile
		\end{algorithmic}
	\end{algorithm}
\subsubsection{Initialization}

Points in the sparse 3D data point cloud generated by SfM, are initialized to 3D Gaussians. The Gaussians are defined by the following variables: position $p$, world space 3D covariance matrix $\Sigma$, opacity $\alpha$, and spherical harmonics coefficient  c (representation of RBG color), given formula: 
\begin{equation}
	G(x)~= e^{-\frac{1}{2}(x)^{T}\Sigma^{-1}(x)}
\end{equation}

\subsubsection{Optimization}

Initially, the Gaussians are sparse and not representative, but they are gradually optimized to better represent the scene. To do this, a random camera view $V$ with it's corresponding image $\hat{I}$ is chosen from the training set. A rasterized Gaussian image $I$ is generated by passing the Gaussian means, $\Sigma$, $c$, $\alpha$, and $V$ to a differentiable rasterizer function $Rasterize()$.

 A loss function, shown in equation \eqref{eq:gsplat_loss}, is used to compute the gradients of the two images $\hat{I}$ and $I$.

\begin{equation}
\label{eq:gsplat_loss}
	\mathcal{L} = (1 - \lambda) \mathcal{L}_1 + \lambda \mathcal{L_{\textrm{D-SSIM}}}
\end{equation}

Here, $\mathcal{L}_1$ is the Mean Absolute Error of $\hat{I}$ and $I$, $\mathcal{L}_{D-SSMI}$ is the Difference Structural Similarity Index based loss, and $\lambda$ is a pre-defined weight that adjusts the contribution of $\mathcal{L}_1$ and $\mathcal{L}_{D-SSMI}$ to the final loss $\mathcal {L}$. The parameters of the Gaussians are adjusted accordingly with the Adam optimizer \cite{kerbl20233d}.

 \subsubsection{Adaptive control of Gaussians}
After initialization, an adaptive approach is used to control the number and density of Guassians. Adaptive control refers to automatically adjusting the size and number of Gaussians to optimize the representation of the static 3D scene. The adaptive density control follows the following behaviors (see Fig. 12):  

\begin{itemize}
    \item \textbf{Gaussian Removal: }For every 100 iterations, if the Gaussians are too large in the 3D space or have opacity under a defined threshold of opacity $\epsilon_\alpha$(essentially transparent), they are removed.
    
    \item \textbf{Gaussian Duplication: }For regions filled by gaussians that are greater than the defined threshold but is too small, the Gaussians are cloned and moved along their direction to cover the empty space.  This behavior adaptively and gradually increases the number and volume of the Gaussians until the area is well-fitted.
    
    \item \textbf{Gaussian Split: }For regions that are over-reconstructed by large Gaussians (variance is too high), they are split into smaller Gaussians by a factor $\phi$, the original paper used $\phi = 1.6$. 
\end{itemize}
    \begin{figure}[!h]
    	\includegraphics[width=\linewidth]{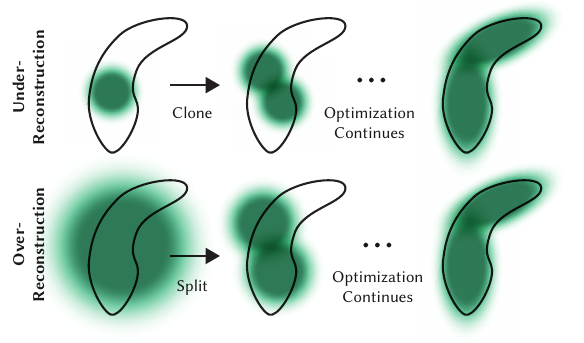}
     	\label{fig:density_control}
    	\caption{
    		Gaussians split in over-reconstructed areas while clone in under-reconstructed areas.
    	}
    \end{figure}

Due to the splitting and duplicating, the number of Gaussians increases. To address this, every Gaussians' opacity $\alpha$ close to zero every $N=3000$ iteration, after the optimization step then increases the $\alpha$ values for the Guassian where this is needed while allowing the unused ones to be removed. 

\subsection{Gradient computation of $\Sigma$ and $\Sigma'$}

Each Gaussian is represented as an ellipsoid in the 3D space, modelled by a covariance matrices $\Sigma$.
\begin{equation}
\label{eq:covariance}
\Sigma=RSS^TR^T
\end{equation}
Where S is the scaling matrix and R is the rotation matrix.

For each camera angle, 3D Gaussians are rasterized to the 2D screen. The 3D covariance matrix $\Sigma'$ is derived using the viewing transformation matrix $W$ and Jacobian $J$, which approximates the projective transformation.

\begin{equation}
\label{eq:sigma_prime}
\Sigma'=JW {\Sigma} W^TJ^T
\end{equation}

$\Sigma'$, a 3x3 matrix, can be simplified by ignoring the elements in the third row and column while retaining the properties of their corresponding planar points representation \cite{zhang2023differentiable}.

To compute the gradient of the 3D space covariance, the chain rule is applied to $\Sigma$ and $\Sigma'$ with reference to their rotation $q$ and scaling $s$. This results in the expressions $\frac{d\Sigma'}{ds} = \frac{d\Sigma'}{d\Sigma}\frac{d\Sigma}{ds}$ and $\frac{d\Sigma'}{dq} = \frac{d\Sigma'}{d\Sigma}\frac{d\Sigma}{dq}$. By substituting $U=JW$ into the equation for $\Sigma'$, we get ${\Sigma'}=U\Sigma U^T$. This equation allows the partial derivative of each element in $\frac{d\Sigma'}{d\Sigma}$ to be represented in terms of $U$:
 
$\frac{\partial \Sigma'}{\partial \Sigma_{ij}} = \left(\begin{smallmatrix}
			U_{1,i}U_{1,j} & U_{1,i} U_{2,j}\\
			U_{1,j} U_{2,i} & U_{2,i}U_{2,j}
		\end{smallmatrix}\right)$

By substituting $M=RS$ into equation \eqref{eq:covariance}, it can then be rewritten as $\Sigma = MM^T$. Using the chain rule, we can derive $\frac{d\Sigma}{ds} = \frac{d\Sigma}{dM}\frac{dM}{ds}$ and $\frac{d\Sigma}{dq} = \frac{d\Sigma}{dM}\frac{dM}{dq}$. This allows us to calculate the scaling gradient at position $(i,j)$ of the covariance matrix with:

$$
\frac{\partial M_{i,j}}{\partial s_k} = \left\{
\begin{array}{lr}
R_{i,k} & \text{if } j = k\\
0 & \text{otherwise}
\end{array}
\right.
$$

For defining derivatives of $M$ with respect to rotation matrix $R$ in terms of quaternion $q$ components, the following formula demonstrating how quaternion components affect $R$ is involved:

	\begin{equation}
     \scriptsize
		R(q) = 2\begin{pmatrix}
			\frac{1}{2} - (q_j^2 + q_k^2) & (q_i q_j - q_r q_k) & (q_i q_k + q_r q_j)\\
			(q_i q_j + q_r q_k) & \frac{1}{2} - (q_i^2 + q_k^2) & (q_j q_k - q_r q_i)\\
			(q_i q_k - q_r q_j) & (q_j q_k + q_r q_i) & \frac{1}{2} - (q_i^2 + q_j^2)
		\end{pmatrix}
		\label{eq:quat}
	\end{equation}

And therefore the gradient $\frac{\partial M}{\partial q_x}$ for 4 components of quaternion $r,i,j,k$ can be calculated as follow:
	\begin{equation}
    \scriptsize
    \begin{aligned}
        &\frac{\partial M}{\partial q_r} = 2 \left(\begin{smallmatrix}
            0 & -s_y q_k & s_z q_j\\
            s_x q_k & 0 & -s_z q_i\\
            -s_x q_j & s_y q_i & 0
        \end{smallmatrix}\right), 
        &\frac{\partial M}{\partial q_i} = 2\left(\begin{smallmatrix}
            0 & s_y q_j & s_z q_k\\
            s_x q_j & -2 s_y q_i & -s_z q_r\\
            s_x q_k & s_y q_r & -2 s_z q_i
        \end{smallmatrix}\right)
        \\
        &\frac{\partial M}{\partial q_j} = 2\left(\begin{smallmatrix}
            -2 s_x q_j & s_y q_i & s_z q_r\\
            s_x q_i & 0 & s_z q_k\\
            -s_x q_r & s_y q_k & -2s_z q_j
        \end{smallmatrix}\right),
        &\frac{\partial M}{\partial q_k} = 2\left(\begin{smallmatrix}
            -2 s_x q_k & -s_y q_r & s_z q_i\\
            s_x q_r & -2s_y q_k & s_z q_j\\
            s_x q_i & s_y q_j & 0
        \end{smallmatrix}\right)
    \end{aligned}
\end{equation}

\subsection{Tile-based rasterizer for real-time rendering}

A technique called Tile-based Rasterizer is used to quickly render the 3D model constructed by the Gaussians (see Fig 13). This approach first uses a given view angle $V$ of the camera and its position $p$ to filter out the Gaussians that are not contributing to the view frustum. In this way only the useful Gaussians are involved in the $\alpha$-blending, improving the rendering efficiency.  

\begin{figure}[!h]
    \includegraphics[width=\linewidth]{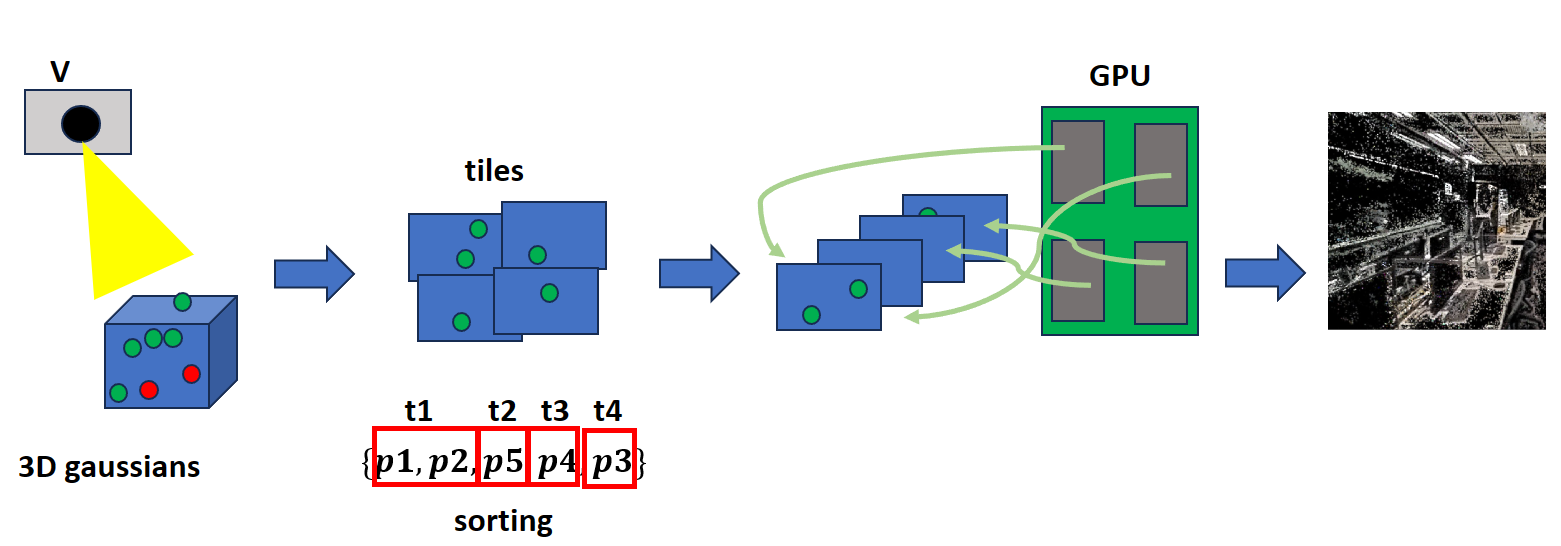}
    \caption{Overview of Tile-based rasterize areas, where green Gaussians are contributing to the view frustum, and red Gaussians are not, given camera angle $V$.}
    \label{fig:density_control}
\end{figure}

Instead of sorting all the Gaussian individually for per-pixel $\alpha$-blending, tiles are created to divide the 2D screen into smaller 16x16 sections. An instance key is assigned to each of the Gaussians using their corresponding view-space depth and the tiles they reside with a tile ID. The Gaussians are then sorted according to their instance key using a GPU Radix sort. The Radix sort is capable of handling large sets of data in parallel with GPU threads. The result of the Gaussian sorting can also demonstrate the depth level of tiles.

After sorting, a thread block is assigned to each tile, and the Gaussians are loaded into the corresponding memory. $\alpha$-blending is then performed from front to back on the sorted list of Gaussians onto the 2D scene, using the cumulative color and opacity $\alpha$ of the Gaussians until each pixel reaches a target alpha saturation. This design maximizes computation efficiency by enabling parallelism of both tile rendering and Gaussian loading in the shared memory space. 
	\begin{algorithm}
		\caption{GPU software rasterization of 3D Gaussians\\
			$w$, $h$: width and height of the image to rasterize\\
			$M$, $S$: Gaussian means and covariances in world space\\
			$C$, $A$: Gaussian colors and opacities\\
			$V$: view configuration of current camera}
		\label{alg:rasterize}
		\begin{algorithmic}

			\Function{Rasterize}{$w$, $h$, $M$, $S$, $C$, $A$, $V$}
			
			\State CullGaussian($p$, $V$) \Comment{Frustum Culling}
			\State $M', S'$ $\gets$ ScreenspaceGaussians($M$, $S$, $V$) \Comment{Transform}
			\State $T$ $\gets$ CreateTiles($w$, $h$)
			\State $L$, $K$ $\gets$ DuplicateWithKeys($M'$, $T$) \Comment{Indices and Keys}
			\State SortByKeys($K$, $L$)							\Comment{Globally Sort}
			\State $R$ $\gets$ IdentifyTileRanges($T$, $K$)
			\State $I \gets \mathbf{0}$ \Comment{Init Canvas}
			
			\ForAll{Tiles $t$ $\textbf{in}$ $I$}
			\ForAll{Pixels $i$ $\textbf{in}$ $t$}
			
			\State $r \gets$ GetTileRange($R$, $t$)
			
			\State $I[i] \gets$ BlendInOrder($i$, $L$, $r$, $K$, $M'$, $S'$, $C$, $A$)

			\EndFor
			\EndFor

			\Return $I$
			\EndFunction
			
		\end{algorithmic}
	\end{algorithm}
\subsection{Limitations}

\subsubsection{Large memory bandwidth}
To achieve real-time rendering with high frames per second, a parallel computing approach is used in the $Rasterize()$ function. This involves a large amount of dynamic data loading occurring in the shared memory of each tile during the $\alpha$-blending process. Therefore, a large GPU memory bandwidth is required to support the data traffic.

\subsubsection{Robustness in Sparse Viewpoints}
Gaussians are optimized by taking the gradient compared with the true camera view. However, viewpoints with few or no data points have less data to optimize the Guassians in their region resulting in artifacts and distortions. 
\section{Future Trends}

\subsubsection{Semantic-Driven 3D Reconstruction} Many 3D reconstruction techniques focus on generating 3D models from images. Yet, the integration of text prompts as a guiding factor presents an exciting avenue for future research. For example, the method outlined in the paper "Semantic-Driven 3D Reconstruction from Single Images"\cite{zhang2023facednerf} demonstrates how textual cues can significantly enhance both the precision and contextual relevance of reconstructed models. While, "LGM: Large Multi-View Gaussian Model for High-Resolution 3D Content Creation" \cite{tang2024lgm} demonstrate impressive zero-shot 3D generations from only a text prompt.

\subsubsection{Dynamic 3D scene reconstruction}
The previously mentioned approaches can only use captured information in static scenes to reproduce static 3D models, where any structure change of scene during information capturing will result in misinformation that leads to under-reconstruction in specific areas. To achieve dynamic 3D scene reconstruction, 4D Gaussian Splatting utilizes a set of canonical 3D Gaussians and transforming them through a deformation field at different times, resulting in producing dynamically changing 3D models that can represent the motion of objects over time \cite{wu20234d}.

\subsubsection{Single View 3D reconstruction}
Building on the methodology introduced in Zero 1-to-3, an area that has gained significant traction is Single View 3D reconstruction. Leveraging diffusion models to generate 3D objects from a single image, \cite{tang2024lgm} and \cite{xu2024instantmesh} have demonstrated promising work in this domain.
\bibliographystyle{plain}
\bibliography{ref}

\begin{thebibliography}{10}

\bibitem{ho2020denoising}
Jonathan Ho, Ajay Jain, and Pieter Abbeel.
\newblock Denoising diffusion probabilistic models, 2020.

\bibitem{kerbl20233d}
Bernhard Kerbl, Georgios Kopanas, Thomas Leimkühler, and George Drettakis.
\newblock 3d gaussian splatting for real-time radiance field rendering, 2023.

\bibitem{Kristiadi2016}
Agustinus Kristiadi.
\newblock Kl divergence: Forward vs reverse?
\newblock \url{https://agustinus.kristia.de/techblog/2016/12/21/forward-reverse-kl/}, 2016.
\newblock Accessed: 2024-04-22.

\bibitem{lee_ddpm_video}
Hung-yi Lee.
\newblock Forward process of ddpm, April 2023.
\newblock https://www.youtube.com/watch?v=ifCDXFdeaaM\&t=608.

\bibitem{liu2023zero1to3}
Ruoshi Liu, Rundi Wu, Basile~Van Hoorick, Pavel Tokmakov, Sergey Zakharov, and Carl Vondrick.
\newblock Zero-1-to-3: Zero-shot one image to 3d object, 2023.

\bibitem{Lombardi:2019}
Stephen Lombardi, Tomas Simon, Jason Saragih, Gabriel Schwartz, Andreas Lehrmann, and Yaser Sheikh.
\newblock Neural volumes: Learning dynamic renderable volumes from images.
\newblock {\em ACM Trans. Graph.}, 38(4):65:1--65:14, July 2019.

\bibitem{martinbrualla2021nerf}
Ricardo Martin-Brualla, Noha Radwan, Mehdi S.~M. Sajjadi, Jonathan~T. Barron, Alexey Dosovitskiy, and Daniel Duckworth.
\newblock Nerf in the wild: Neural radiance fields for unconstrained photo collections, 2021.

\bibitem{mescheder2019occupancy}
Lars Mescheder, Michael Oechsle, Michael Niemeyer, Sebastian Nowozin, and Andreas Geiger.
\newblock Occupancy networks: Learning 3d reconstruction in function space, 2019.

\bibitem{mildenhall2020nerf}
Ben Mildenhall, Pratul~P. Srinivasan, Matthew Tancik, Jonathan~T. Barron, Ravi Ramamoorthi, and Ren Ng.
\newblock Nerf: Representing scenes as neural radiance fields for view synthesis, 2020.

\bibitem{M_ller_2022}
Thomas Müller, Alex Evans, Christoph Schied, and Alexander Keller.
\newblock Instant neural graphics primitives with a multiresolution hash encoding.
\newblock {\em ACM Transactions on Graphics}, 41(4):1–15, July 2022.

\bibitem{park2019deepsdf}
Jeong~Joon Park, Peter Florence, Julian Straub, Richard Newcombe, and Steven Lovegrove.
\newblock Deepsdf: Learning continuous signed distance functions for shape representation, 2019.

\bibitem{Soft3DReconstruction}
Eric Penner and Li~Zhang.
\newblock Soft 3d reconstruction for view synthesis.
\newblock 36(6), 2017.

\bibitem{rombach2022highresolution}
Robin Rombach, Andreas Blattmann, Dominik Lorenz, Patrick Esser, and Björn Ommer.
\newblock High-resolution image synthesis with latent diffusion models, 2022.

\bibitem{tang2024lgm}
Jiaxiang Tang, Zhaoxi Chen, Xiaokang Chen, Tengfei Wang, Gang Zeng, and Ziwei Liu.
\newblock Lgm: Large multi-view gaussian model for high-resolution 3d content creation.
\newblock {\em arXiv preprint arXiv:2402.05054}, 2024.

\bibitem{tremblay2022rtmv}
Jonathan Tremblay, Moustafa Meshry, Alex Evans, Jan Kautz, Alexander Keller, Sameh Khamis, Thomas Müller, Charles Loop, Nathan Morrical, Koki Nagano, Towaki Takikawa, and Stan Birchfield.
\newblock Rtmv: A ray-traced multi-view synthetic dataset for novel view synthesis, 2022.

\bibitem{wu20234d}
Guanjun Wu, Taoran Yi, Jiemin Fang, Lingxi Xie, Xiaopeng Zhang, Wei Wei, Wenyu Liu, Qi~Tian, and Xinggang Wang.
\newblock 4d gaussian splatting for real-time dynamic scene rendering, 2023.

\bibitem{xu2024instantmesh}
Jiale Xu, Weihao Cheng, Yiming Gao, Xintao Wang, Shenghua Gao, and Ying Shan.
\newblock Instantmesh: Efficient 3d mesh generation from a single image with sparse-view large reconstruction models, 2024.

\bibitem{zhang2023facednerf}
Hao Zhang, Yanbo Xu, Tianyuan Dai, Yu-Wing Tai, and Chi-Keung Tang.
\newblock Facednerf: Semantics-driven face reconstruction, prompt editing and relighting with diffusion models, 2023.

\bibitem{zhang2023differentiable}
Qiang Zhang, Seung-Hwan Baek, Szymon Rusinkiewicz, and Felix Heide.
\newblock Differentiable point-based radiance fields for efficient view synthesis, 2023.

\end{thebibliography}

\end{document}